\documentclass{bmvc2k}
\usepackage{amssymb}
\usepackage{amsmath}
\usepackage{algorithm}
\usepackage{algorithmicx}
\usepackage{bm}
\usepackage{graphicx}
\usepackage{multicol,multirow}
\usepackage{color}
\usepackage{epstopdf}
\usepackage{hyperref}
\usepackage{marvosym}
\usepackage{lipsum}

\newcommand{\xu}[1]{\textcolor{black}{#1}}
\makeatletter
  \newcommand\figcaption{\def\@captype{figure}\caption}
  \newcommand\tabcaption{\def\@captype{table}\caption}
\makeatother

\title{POEM: 1-bit Point-wise Operations based on Expectation-Maximization for Efficient Point Cloud Processing}

\addauthor{Sheng Xu}{shengxu@buaa.edu.cn}{1}
\addauthor{Yanjing Li}{yanjingli@buaa.edu.cn}{1}
\addauthor{Junhe Zhao}{jhzhao@buaa.edu.cn}{1}
\addauthor{Baochang Zhang}{bczhang@buaa.edu.cn}{1}
\addauthor{Guodong Guo}{guoguodong01@baidu.com}{2}

\addinstitution{
 Beihang University,\\
 Beijing, China
}
\addinstitution{
 National Engineering Laboratory for Deep Learning Technology and Application,\\ Institute of Deep Learning,\\
 Baidu Research,\\
 Beijing, China
}

\runninghead{Xu \MakeLowercase{\rm and} Li \MakeLowercase{\rm et} \MakeLowercase{\rm al}.}{POEM: 1-bit Point-wise Operations based on E-M}

\begin{document}

\maketitle
\begin{abstract}
Real-time point cloud processing is fundamental for lots of computer vision tasks, while still  challenged by the computational problem  on resource-limited edge devices. To address this issue, we  implement  XNOR-Net-based binary neural networks (BNNs) for an efficient point cloud processing, but its performance is severely suffered due to two main drawbacks, \xu{Gaussian}-distributed weights and non-learnable scale factor. In this paper, we introduce  point-wise operations based on  Expectation-Maximization (POEM) into BNNs  for efficient  point cloud processing. The EM algorithm can efficiently constrain weights for a robust bi-modal distribution. We lead a well-designed reconstruction loss  to calculate learnable scale factors to enhance the representation capacity of 1-bit fully-connected (Bi-FC) layers. Extensive experiments demonstrate that our POEM surpasses existing \xu{the state-of-the-art binary point cloud networks} by a significant margin, up to 6.7\%.
\end{abstract}

\section{Introduction}
\label{sec:intro}
Compared with traditional 2D images, 3D data provides an opportunity to understand the surrounding environment for machines better. With the advancement of deep neural networks (DNNs) directly processing raw point clouds, great success has been achieved in  PointNet \cite{qi2017pointnet}, PointNet++ \cite{qi2017pointnet++} and DGCNN \cite{wang2019dynamic}. However, existing methods are inefficient for real applications that require real-time inference and fast response, such as autonomous driving and augmented reality. Their deployed environments are often resource-constrained edge devices. To address the challenge,  Grid-GCN \cite{xu2020grid}, RandLA-Net \cite{hu2020randla}, and PointVoxel \cite{liu2019point}, have been introduced for efficient point cloud processing using 
DNNs. While significant speedup and memory footprint reduction have been achieved, these works still rely on expensive floating-point operations, leaving room for further optimization of the performance from the model quantization perspective.  Binarized neural network (BNNs) \cite{courbariaux2015binaryconnect,rastegari2016xnor,liu2018bi,gu2019projection,gu2019bayesian,liu2020reactnet,wang2020bidet,xu2020amplitude,xu2020layer} compress weights and activations of DNNs into a single bit, which can decrease the storage requirements by $32\times$ and computation cost by up to $58\times$ \cite{rastegari2016xnor}. However, using network binarization for point cloud processing tasks remains largely unexplored. 

\begin{figure*}
	\centering
	\includegraphics[scale=.35]{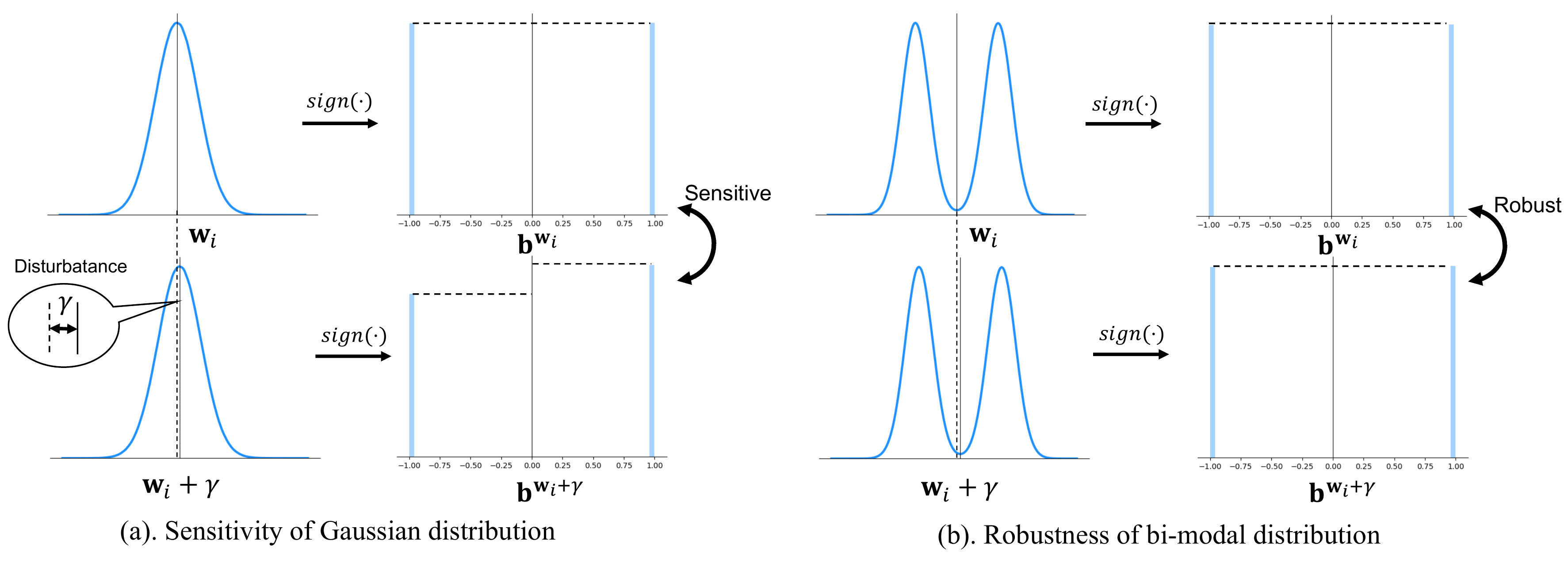}
	\caption{Subfigure (a) and (b) illustrates the robustness of \xu{Gaussian} distribution and bi-modal distribution. From left to right in each subfigure, we plot the distribution of the unbinarized weights ${\bf w}_i$ and the binarized weights ${\bf b}^{{\bf w}_i}$.} 
	\label{xnor}
\end{figure*}

In this paper, we first implement a baseline, XNOR-Net-based \cite{rastegari2016xnor} 1-bit point cloud network, which shows that the performance drop is mainly caused by two drawbacks. First, layer-wise weights of XNOR-Net roughly follow a \xu{Gaussian} distribution with a mean value around 0. However, such a distribution is subjected to the disturbance aroused by the noise containing in the raw point cloud data \cite{hermosilla2019total}. {As a result,  the \xu{Gaussian}-distributed weight (around 0) will accordingly change its sign, {\em i.e.}, the binarization result will be changed dramatically.} This explains why the baseline network is ineffective to process the point cloud data and achieves a worse convergence, as shown in Figure~\ref{xnor} (a). {In contrast, bi-modal distribution will gain more robustness against the noise.} Second, XNOR-Net fails to adapt itself to the characteristics of cloud data, when computing the scale factor using a non-learning method.

To address these issues, we introduce 1-bit point-wise operations based on Expectation-Maximization (POEM) to efficiently process the point cloud data. First, we exploit  Expectati-
on-Maximization (EM) \cite{moon1996expectation} to constrain the weights into a bi-modal distribution, which can be more robust to disturbances caused by the noise containing in the raw point cloud data \cite{hermosilla2019total}, as shown in Figure~\ref{xnor} (b). We  also introduce a learnable and adaptive scale factor for every 1-bit layer to enhance the feature representation capacity of our binarized networks. Finally, we lead a powerful 1-bit network for point cloud processing, which can well reconstruct real-valued counterparts' amplitude via a new learning-based method. Our contributions are summarized as follows:
\begin{itemize}
    \item  We introduce a new binarization approach of point-wise operations based on Expectati-
    on-Maximization (POEM), which  can efficiently binarize network weights and activations for point cloud processing.
    \item We achieve a learnable scale factor to modulate the amplitude of real-valued weights in an end-to-end manner, which can significantly improve  the representation ability of  binarized networks.
    \item Our methods are generic and can be readily extendable to mainstream point cloud feature extractors. Extensive experiments on multiple fundamental point cloud tasks demonstrate the superiority of our POEM. For example, the 1-bit PointNet mounted by our POEM achieves 90.2\% overall accuracy on the ModelNet40 dataset, which is even 0.9\% higher than real-valued counterparts and promotes the state-of-the-arts.
 \end{itemize}
\begin{figure*}
	\centering
	\includegraphics[scale=.3]{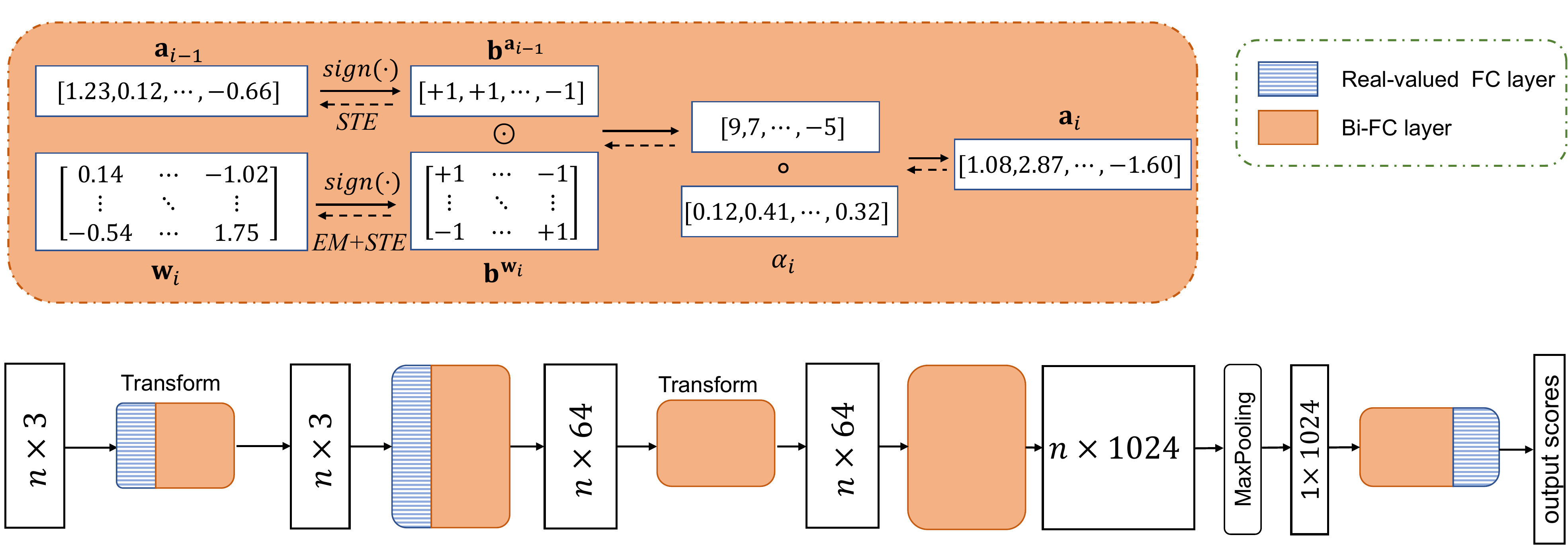}
	\caption{Outline of the 1-bit PointNet obtained by our POEM on the classification task. EM denotes Expectation-Maximization algorithm, and STE denotes Straight-Through-Estimator.} 
	\label{main}
\end{figure*}

\section{Learning networks via POEM}

This section elaborates our proposed POEM method, including the binarization framework, the supervision for learning a scale factor, and the optimization towards robust weights distribution through the EM method.

 \subsection{Binarization Framework of POEM}
Our POEM framework is  shown in Figure~\ref{main}. We extend the binarization process from 2D convolution (XNOR-Net) to fully-connected layers (FCs) for feature extraction, termed 1-bit fully-connected (Bi-FC) layers, based on extremely efficient bit-wise operations (XNOR and Bit-count) via the lightweight binary weight and activation.

Given a conventional FC layer, we denote ${\bf w}_i\in \mathbb{R}_{C_{i}\times C_{i-1}}$ and ${\bf a}_i\in \mathbb{R}_{C_i}$ as its weights and features in the $i$-th layer. $C_i$ represents the number of output channels of $i$-th layer. We then have
${\bf a}_i = {\bf a}_{i-1} \otimes {\bf w}_{i}$,
where $\otimes$ denotes full-precision multiplication. 
As mentioned above, the BNN model aims to binarize $ {\bf w}_{i}$ and $ {\bf a}_{i}$ into ${\bf b}^{{\bf w}_i}\in \mathbb{B}_{m_i}$ and ${\bf b}^{{\bf a}_i}\in \mathbb{B}_{C_i}$ in this paper respectively, where $\mathbb{B}$ denotes discrete set $\{-1, +1\}$ for simplicity. Then, we apply XNOR and Bit-count operations to replace full-precision operations. Following \cite{rastegari2016xnor}, the forward process of the BNN is defined as
\begin{equation}
{\bf a}_i = {\bf b}^{{\bf a}_{i-1}} \odot {\bf b}^{{\bf w}_i},
\label{8}
\end{equation}
where $\odot$ represents efficient XNOR and Bit-count operations. Based on XNOR-Net \cite{rastegari2016xnor}, we introduce a learnable channel-wise scale factor to modulate the amplitude of real-valued convolution. Considering the Batch Normalization (BN) and activation layers, the forward process is formulated as
\begin{equation}
{\bf b}^{{\bf a}_i} = sign(\Phi({\alpha}_i\circ{\bf b}^{{\bf a}_{i-1}} \odot {\bf b}^{{\bf w}_i})),
\label{9}
\end{equation}
where we divide the data flow in POEM into units for detailed discussions. In POEM, the original output feature ${\bf a}_i$ is first scaled by a channel-wise scale factor (vector) ${\alpha}_i\in \mathbb{R}_{C_i}$ to modulate the amplitude of full-precision counterparts. It then enters $\Phi(\cdot)$, which represents a composite function built by stacking several layers, {\em e.g.}, BN layer, non-linear activation layer, and max-pooling layer. And then, the output is binarized to obtain the binary activations ${\bf b}^{{\bf a}_i}\in \mathbb{B}_{C_i}$, via sign function. ${\rm sign(\cdot)}$ denotes the sign function which returns $+1$ if the input is greater than zeros, and $-1$ otherwise. Then, the 1-bit activation ${\bf b}^{{\bf a}_i}$ can be used for the efficient XNOR and Bit-count of $(i\!\!+\!\!1)$-th layer.
\subsection{Supervision for POEM}
To constrain the Bi-FC to have binarized weights with similar amplitudes as the real-valued counterparts, we introduce a new loss function in our supervision for POEM. We consider that unbinarized weights should be reconstructed based on binarized weights, and define the reconstruction loss as
\begin{equation}
L_R = \frac{1}{2}\|{{\bf w}_i}-{\alpha}_i \circ {\bf b}^{{\bf w}_i}\|_2^2,
\label{10}
\end{equation}
where $L_R$ is the reconstruction loss. Considering the impact of ${\alpha}_i$ on the output of the layer, we define the learning objective of our POEM as 
\begin{equation}
\mathop{\arg\min}_{\{{\bf w}_i, {\alpha}_i\},\forall i\in N} L_S({\bf w}_i, {\alpha}_i)+\lambda L_R({\bf w}_i, {\alpha}_i),
\label{11}
\end{equation}
where $N$ denotes the number of layers in the network. \xu{$L_S$ is the cross entropy, denoting learning from the ground truth.} And $\lambda$ is a hyper-parameter. Different from binarization methods (such as XNOR-Net \cite{rastegari2016xnor} and Bi-Real Net \cite{liu2018bi}) where only the reconstruction loss is considered in the weight calculation. {By fine-tuning the value of $\lambda$, our proposed POEM can achieve much better performance than XNOR-Net, which shows the effectiveness of combined loss against only softmax loss.} 

\begin{figure*}
	\centering
	\includegraphics[scale=.38]{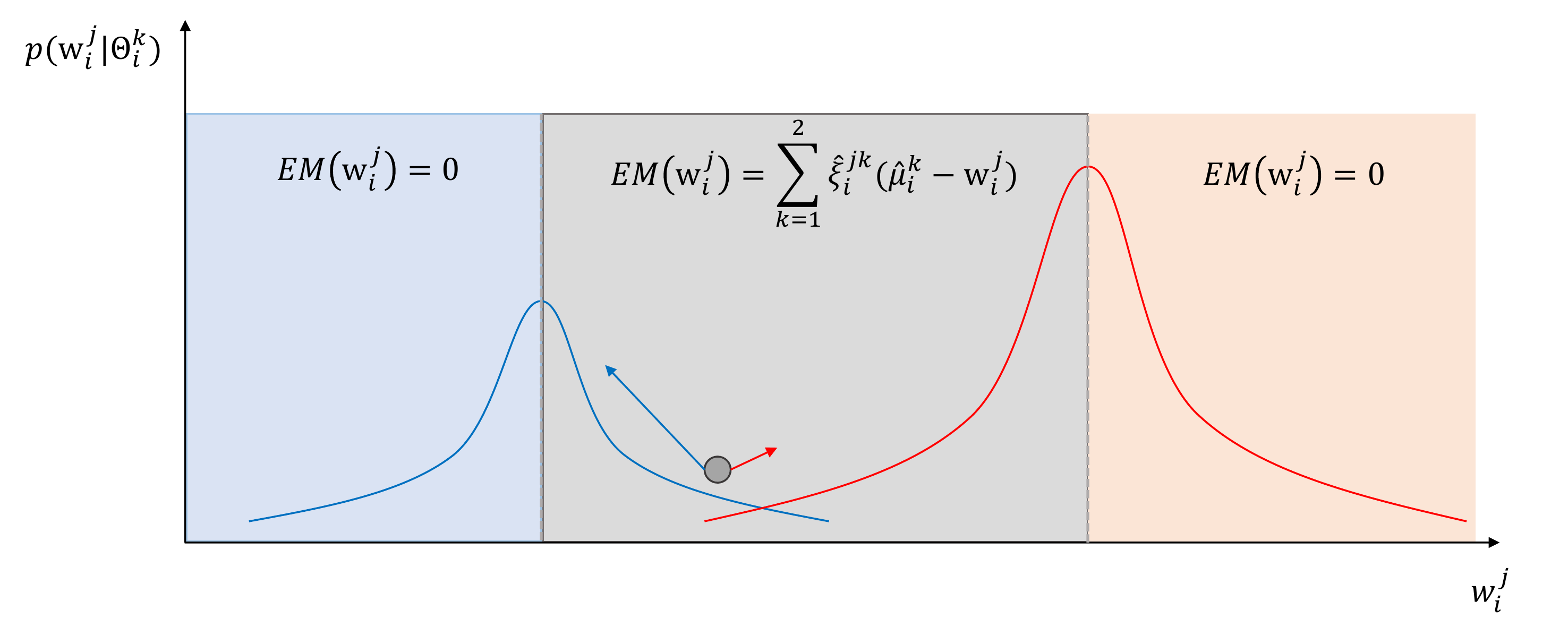}
	\caption{Illustration of training ${\rm w}^j_i$ via Expectation-Maximization. For the ones in the grey area (distribution not transparent), we apply $EM(\cdot)$ to constrain it to converge to a specific distribution.} 
	\label{em}
\end{figure*}

\subsection{Optimization for POEM}
In each Bi-FC layer, POEM sequentially update unbinarized weights ${\bf w}_i$ and scale factor ${\alpha}_i$.

\noindent\textbf{Updating ${\bf w}_i$ via Expectation-Maximization:}
Under a binarization framework, the conventional back propagation process without necessary constraint will result in a \xu{Gaussian} distribution of ${\bf w}_i$, which degrades the robustness of Bi-FCs. Our POEM takes another learning objective as
\begin{equation}
\xu{\mathop{\arg\min}_{{\bf w}_{i,j}} {\bf b}^{{\bf w}_{i,j}}-{\bf b}^{{\bf w}_{i,j}+{\gamma}}, }
\label{13}
\end{equation}
where $\gamma$ denotes any disturbance caused by the noise containing in the raw point cloud data and $j=1,\cdots,C_i$ is the channel index. To learn Bi-FCs capable of overcoming this obstacle, we introduce EM algorithm into the updating of ${\bf w}_{i,j}$. First, we assume the ideal distribution of ${\bf w}_{i,j}$ should be a bi-modal one.

\newtheorem{assumption}{Assumption}[section]
\begin{assumption}
\xu{For every channel of the $i$-th 1-bit layer, i.e., $\forall{\bf w}_{i,j}\in {\bf w}_i$, it can be constrained to follow a Gaussian Mixture Model (GMM).}
\end{assumption}

Based on our assumption, for \xu{the $j$-th channel ${\bf w}_{i,j}$} we formulate the ideal bi-modal distribution as
\begin{equation}
\xu{\mathcal{P}({\bf w}_{i,j}|\bm{\Theta}_{i,j})=\beta_{i,j}^k\sum^{2}_{k=1}p({\bf w}_{i,j}|\Theta_{i,j}^k),}
\label{13}
\end{equation}
where the number of distributions is set as $2$ in this paper. $\Theta^k_{i,j}=\{\mu_{i,j}^k, \sigma_{i,j}^k\}$ denotes the parameters of the $k$-th distribution, {\em i.e.}, $\mu_{i,j}^k$ denotes the mean value and $\sigma_{i,j}^k$ denotes the variance respectively. To solve GMM with the observed data ${\bf w}_i$, {\em i.e.}, the ensemble of weights at the $i$-th layer.

We introduce the hidden variable $\xi_{i,j}^{nk}$ to formulate the maximum likelihood estimation (MLE) of GMM as
\begin{equation}
\xu{\xi_{i,j}^{jk}= \left\{
\begin{aligned}
&1,\ \ {\rm w}^n_{i,j}\in p^k_{i,j} \\
&0,\ \ \rm{else}
\end{aligned}
\right.,}
\label{14}
\end{equation}
where $n=1,\cdots,C_{i-1}$ is the input channel index of the $i$-th layer. $\xi_{i,j}^{jk}$ is the hidden variable describing the affiliation of ${\rm w}^n_{i,j}$ and $p^k_{i,j}$ (simplified denotation of $p({\bf w}_{i,j}|\Theta_{i,j}^k)$). We then define the likelihood function $\mathcal{P}({\rm w}^n_{i,j}, \xi_{i,j}^{nk}|\Theta^k_{i,j})$ as 

\begin{equation}
\xu{\mathcal{P}({\rm w}^n_{i,j}, \xi^{nk}_{i,j}|{\Theta}^k_{i,j})=\prod_{k=1}^{2}(\beta_{i,j}^k)^{|p_{i,j}^k|}\prod_{n=1}^{C_{i-1}}\left\{\frac{1}{\Omega}f({\rm w}_{i,j}^n, \mu_{i,j}^k, \sigma^k_{i,j})\right\}^{\xi_{i,j}^{nk}},}
\end{equation}
where $\Omega\!=\!\sqrt{2\pi|\sigma^k_{i,j}|}$, $|p_{i,j}^k|\!=\!\sum^{C_{i-1}}_{n=1}\xi_{i,j}^{nk}$, and $C_{i-1}\!=\!\sum^2_{k=1}|p_{i,j}^k|$. And $f({\rm w}^n_{i,j}, \mu_{i,j}^k, \sigma^k_{i,j})$ is defined as 
\begin{equation}
\xu{f({\rm w}^n_{i,j}, \mu_{i,j}^k, \sigma^k_{i,j})={\rm exp}(-\frac{1}{2\sigma^k_{i,j}}({\rm w}_{i,j}^n-\mu_{i,j}^k)^2).}
\end{equation}
Hence, for every single weight ${\rm w}^n_{i,j}$, $\xi^{nk}_{i,j}$ can be computed by maximizing the likelihood as 
\begin{equation}
\xu{\mathop{\max}_{\xi_{i,j}^{nk}, \forall n, k}\;\;\mathbb{E}\left[{\rm log}\;\mathcal{P}({\rm w}^n_{i,j},\xi_{i,j}^{nk}|\Theta_{i,j}^k)|{\rm w}^n_{i,j},\Theta_{i,j}^k\right],}
\end{equation}
where $\mathbb{E}[\cdot]$ represents the estimation. Hence, the maximum likelihood estimation {$\hat{\xi}_{i,j}^{nk}$} is calculated as
\begin{equation}
\xu{\hat{\xi}_{i,j}^{nk}=\frac{\beta^k_{i,j} p({\rm w}^n_{i,j}|\Theta^k_{i,j})}{\sum^{2}_{k=1}\beta^k_{i,j} p({\rm w}^n_{i,j}|\Theta^k_{i,j})}.}
\end{equation}

After the expectation step, we conduct the maximization step to compute $\Theta^k_{i,j}$ as 
\begin{equation}
\xu{(\hat{\mu}^{k}_{{i,j}},\hat{\sigma}^{k}_{{i,j}},\hat{\beta}^{k}_{{i,j}})=(\frac{\sum^{C_{i-1}}_{n=1}\hat{\xi}^{nk}_{{i,j}}{\rm w}^{n}_{{i,j}}}{\sum^{C_{i-1}}_{n=1}\hat{\xi}_{{i,j}}^{nk}},\frac{\sum^{C_{i-1}}_{n=1}\hat{\xi}^{nk}_{{i,j}}({\rm w}^{n}_{{i,j}}-\hat{\mu}^{k}_{{i,j}})^2}{\sum^{C_{i-1}}_{n=1}\hat{\xi}_{{i,j}}^{nk}},\frac{\sum^{C_{i-1}}_{n=1}\hat{\xi}^{nk}_{{i,j}}}{C_{i-1}}).}
\end{equation}
Then, we optimize ${\rm w}^n_{{i,j}}$ as
\begin{equation}
\xu{\delta_{{\rm w}^n_{{i,j}}}=\frac{\partial L_S}{\partial {\rm w}^n_{i,j}}+\lambda\frac{\partial L_R}{\partial {\rm w}^n_{i,j}} + \tau EM({\rm w}^n_{i,j}),}
\label{24}
\end{equation}
where $\tau$ is hyper-parameter to control the proportion of Expectation-Maximization operator $EM({\rm w}^n_{i,j})$. $EM({\rm w}^n_{i,j})$ is defined as 
\begin{equation}
\begin{aligned}
\xu{EM({w}^n_{i,j})=\left\{
\begin{array}{rcl}
\sum^2_{k=1}\hat{\xi}^{jk}_{i}({\hat{\mu}^{k}_{{i,j}}-{\rm w}^n_{i,j}}),& {\hat{\mu}^{1}_{{i,j}} < {\rm w}^n_{i,j}  < \hat{\mu}^{2}_{{i,j}}}\\
0,& {\rm else}\\
\end{array} \right..}
\end{aligned}
\label{25}
\end{equation}
And further we have
\begin{equation}
\frac{\partial L_R}{\partial {\bf w}_i} = ({{\bf w}_i}-{\alpha}_i \circ {\bf b}^{{\bf w}_i})\circ{\alpha}_i.
\end{equation}

\noindent\textbf{Updating ${\alpha}_i$:}
We further update the scale factor ${\alpha}_i$ with ${\bf w}_i$ fixed. $\delta_{{\alpha}_i}$ is defined as the gradient of ${\alpha}_i$, and we have

\begin{equation}
\delta_{{\alpha}_i}=\frac{\partial L_S}{\partial {\alpha}_i}+\lambda\frac{\partial L_R}{\partial {\alpha}_i}.
\label{27}
\end{equation}
The gradient derived from softmax loss can be easily calculated according to back propagation. Base on Eq.~\ref{10}, we have
\begin{equation}
\frac{\partial L_R}{\partial {\alpha}_i}={({\bf w}_i-{\alpha}_i \circ {\bf b}^{{\bf w}_i})\cdot {\bf b}^{{\bf w}_i}}.
\label{29}
\end{equation}

The above derivations show that POEM is learnable with the BP algorithm  based on a simple and effective reconstruction loss function. Moreover, we introduce EM to optimize unbinarized weights, which further constrain them  to formulate a bi-modal distribution. We describe our algorithm in supplementary materials.

\section{Implementation and Experiments}

\begin{figure}[t]
    \begin{minipage}[t]{0.55\textwidth}
    \vspace{-31.5mm}
    \centering
    \renewcommand\arraystretch{0.9}
    \setlength{\tabcolsep}{1mm}{\begin{tabular}{|c|c|c|c|c|c|}
    \hline
    \multicolumn{2}{|c|}{{\begin{tabular}[c]{@{}c@{}}1-bit\\ PointNet\end{tabular}}} & \multicolumn{4}{c|}{$\lambda$}         \\ \cline{3-6} 
    \multicolumn{2}{|c|}{}                                                                          & $1\times10^{-3}$ & $1\times10^{-4}$        & $1\times10^{-5}$ & $0$  \\ \hline
    \multirow{4}{*}{$\tau$}                                 & $1\times10^{-2}$                                & 89.3   & 89.0          & 86.3   & 81.9 \\ \cline{2-6} 
                                                            & $1\times10^{-3}$                                & 88.3   & \textbf{90.2} & 87.9   & 82.5 \\ \cline{2-6} 
                                                            & $1\times10^{-4}$                                & 86.5   & 87.1          & 85.5   & 81.4 \\ \cline{2-6} 
                                                            & $0$                                   & 82.7   & 85.3          & 83.7   & 80.1 \\ \hline
    \end{tabular}}
    \vspace{4mm}
	\tabcaption{Ablation study on hyper-parameter $\lambda$ and $\tau$. We vary $\lambda$ from $1\!\times\!10^{-3}$ to $0$ and $\tau$ from $1\!\times\!10^{-2}$ to $0$, respectively.}
	\label{hyper-parameter}
	\end{minipage}
	\hspace{3mm}
	\begin{minipage}[t]{0.4\textwidth}
	\centering
	\includegraphics[scale=.3]{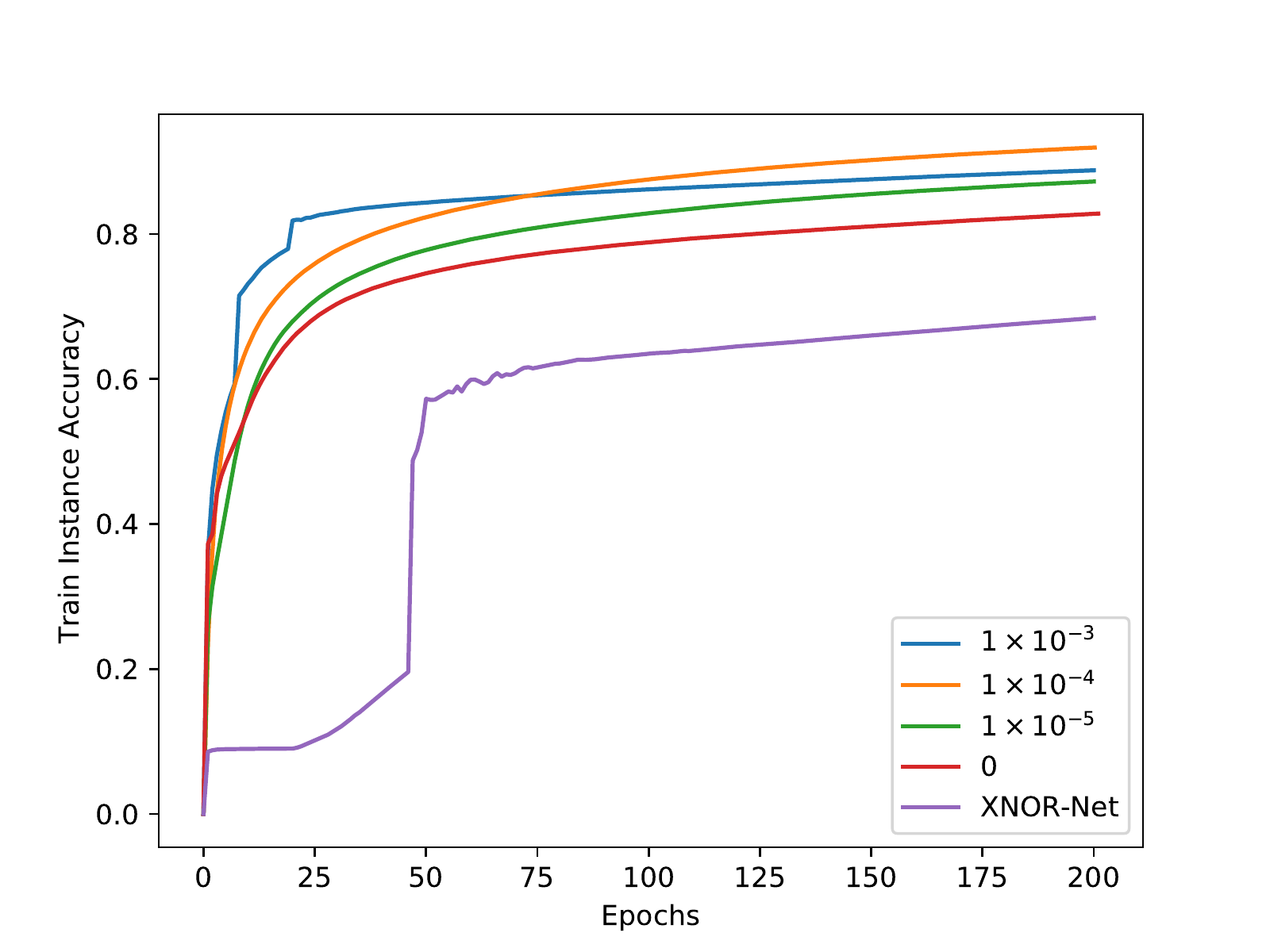}
	\vspace{-3mm}
	\figcaption{Training accuracies of POEM ($\tau=1\times10^{-3}$) with different $\lambda$ and XNOR-Net.}
	\centering
	\label{train_acc}
	\end{minipage}
	\vspace{-3mm}
\end{figure}
In this section, we conduct extensive experiments to validate the effectiveness of our proposed POEM for efficient learning on point clouds. We first ablate our method and demonstrate the contributions of our work on the most fundamental tasks: classification on ModelNet40 \cite{wu20153d}. Moreover, we implement our POEM on mainstream models on three tasks, {\em i.e.}, classification on ModelNet40 \cite{wu20153d}, part segmentation on ShapeNet Parts \cite{chang2015shapenet}, and semantic segmentation on S3DIS \cite{armeni20163d}. We compare POEM with existing binarization methods where our designs stand out. 
\subsection{Datasets and Implementation Details}
\noindent\textbf{Datasets:} ModelNet40 \cite{wu20153d} is used for classification. The ModelNet40 dataset is the most frequently used datasets for shape classification. ModelNet is a popular benchmark for point cloud classification. It contains $12,311$ CAD models from $40$ representative classes of objects.

We employ ShapeNet Parts \cite{chang2015shapenet} for part segmentation. ShapeNet contains $16,881$ shapes from $16$ categories, $2,048$ points are sampled from each training shape. Each shape is split into two to five parts depending on the category, making up to $50$ parts in total.

For the semantic segmentation, S3DIS \cite{armeni20163d} is employed. S3DIS includes 3D scan point clouds for 6 indoor areas, including 272 rooms in total, and each point belongs to one of 13 semantic categories.

\noindent\textbf{Implementation Details:} We evaluate POEM on three mainstream models, including PointNet \cite{qi2017pointnet}, PointNet++ \cite{qi2017pointnet++} and DGCNN \cite{wang2019dynamic}, on three main point cloud tasks, {\em i.e.}, classification, part segmentation and semantice segmentation. In our experiments, 4 NVIDIA GeForce TITAN V GPUs are used. 

On the classification task, 1-bit PointNet is built by binarizing the full-precision PointNet via POEM. All fully-connected layers in PointNet except the first and last one are binarized to the Bi-FC layer, and we select PReLU \cite{he2015delving} instead of ReLU as the activation function when binarizing the activation before the next Bi-FC layer. We also extend this binarization setting to other tasks. We also provide our PointNet baseline under this setting. For other 1-bit networks, we also follow this implementation. In all tables, we use the bold typeface to denote the best result.

For the part segmentation task, we follow the convention \cite{qi2017pointnet} to train a model for each of the 16 classes of ShapeNet Parts \cite{chang2015shapenet}. For semantic segmentation task on S3DIS \cite{armeni20163d}, we also follow the same setups an \cite{qi2017pointnet}.

Following PointNet \cite{qi2017pointnet}, we train $200$ epochs, $250$ epochs, $128$ epochs on point cloud classification, part segmentation, semantic segmentation respectively. To stably train the 1-bit networks, we use learning rate 0.001 with Adam and Cosine Annealing learning rate decay for all 1-bit models on all tasks. 

\begin{figure*}
	\centering
	\includegraphics[scale=.23]{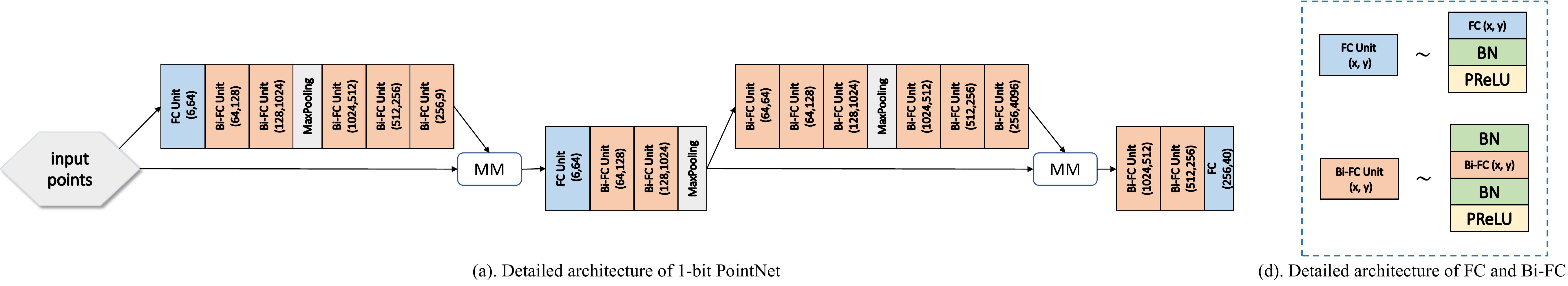}
	\caption{Detailed architecture of 1-bit PointNet implemented by us. MM denotes matrix multiplication in short.} 
	\label{archi}
\end{figure*}
\begin{figure*}[t]
	\centering
	\includegraphics[scale=.28]{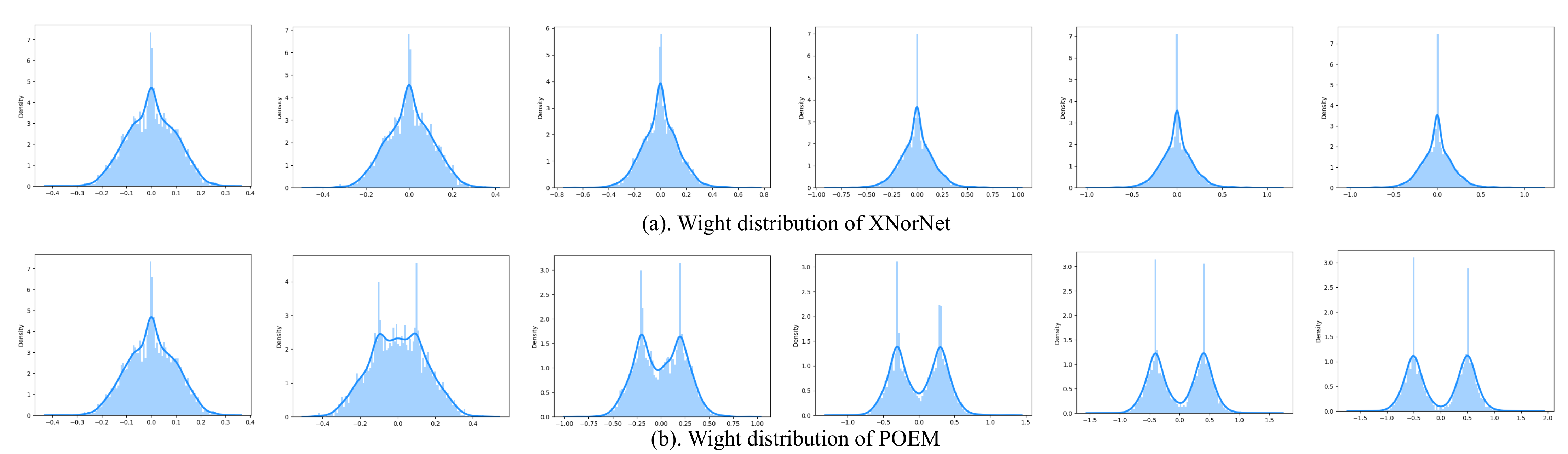}
	\caption{(a) and (b) illustrate the distribution of the unbinarized weights ${\bf w}_i$ of the $6$-th 1-bit layer in 1-bit PointNet backbone when being trained under XNOR-Net and our POEM, respectively. From left to right, we report the weight distribution of initialization, $40$-th, $80$-th, $120$-th, $160$-th, and $200$-th epoch. Our POEM obtains an apparent bi-modal distribution, which is much more robust.}
	\label{distribution}
\end{figure*}
\subsection{Ablation Study}

\noindent\textbf{Hyper-parameter selection:} Hyper-parameters $\lambda$ and $\tau$ in Eq.~\ref{11} and~\ref{24} are related to the reconstruction loss and EM algorithm. The effect of parameters $\lambda$ and $\tau$ are evaluated on ModelNet40 for 1-bit PointNet. The Adam optimization algorithm is used during the training process, with per batch sized as $592$. Using different values of $\lambda$ and $\tau$, the performance of POEM is shown in Table~\ref{hyper-parameter}. In Table~\ref{hyper-parameter}, from left to right lie the overall accuracies (OAs) with different $\lambda$ from $1\!\times\!10^{-3}$ to $0$. 

And the OAs with different $\tau$ from $1\!\times\!10^{-2}$ to $0$ lie from top to bottom. With the decrease of $\lambda$, the OA increases first and then drops dramatically. The same trend is shown when we decrease $\tau$. We get the optimal 1-bit PointNet with POEM with $\{\lambda,\tau\}$ set as $\{1\!\times\!10^{-4},1\!\times\!10^{-3}\}$. Hence, we extend this hyper-parameter set to the other experiments involved in this paper. 

We also set $\tau$ as $1\!\times\!10^{-3}$ and plot the growth curve of training accuracies of POEM with different $\lambda$ and XNOR-Net. As shown in Figure~\ref{train_acc}, 1-bit PointNet obtained by POEM achieves the optimal training accuracy when $\lambda$ is set as $1\!\times\!10^{-4}$. Also, with the EM optimized back propagation, the convergence of weights becomes better than XNOR-Net {(in purple)}, as shown in Figure~\ref{train_acc}.

\noindent\textbf{Evaluating the components of POEM:} In this part, we evaluate every critical part of POEM to show how we compose the novel and effective POEM.
\begin{figure}[t]
    \begin{minipage}[t]{0.55\textwidth}
    \vspace{-21.7mm}
    \centering
    \renewcommand\arraystretch{1.1}
    \setlength{\tabcolsep}{1mm}{\begin{tabular}{|c|c|}
\hline
1-bit PointNet          & OA (\%) \\ \hline
XNOR-Net                 & 81.9             \\
Proposed baseline network                    & 83.1             \\ \hline
Proposed baseline network + PReLU            & 85.0             \\
Proposed baseline network + EM               & 86.2             \\
Proposed baseline network + LSF              & 86.5             \\
\begin{tabular}[c]{@{}c@{}}Proposed baseline network + PReLU \\ + EM + LSF (POEM)\end{tabular}   & 90.2             \\ \hline
Real-valued Counterpart & 89.2         \\ \hline
\end{tabular}}
\vspace{2.7mm}
	\tabcaption{The effects of different components in POEM on the OA. PReLU, EM and LSF denote components of proposed baseline network.}
	\label{component}
	\end{minipage}
	\hspace{1mm}
	\begin{minipage}[t]{0.45\textwidth}
	\centering
	\footnotesize
	\renewcommand\arraystretch{0.89}
\setlength{\tabcolsep}{1mm}{\begin{tabular}{|c|c|c|c|}
\hline
Model                       & Method      & W/A (bit)                  & OA (\%)     \\ \hline
\multirow{5}{*}{PointNet}   & Real-valued & 32/32                & 89.2          \\ \cline{2-4} 
                            & XNOR-Net     & \multirow{4}{*}{1/1} & 81.9          \\
                            & Bi-Real Net &                      & 77.5          \\
                            & BiPointNet  &                      & 86.4          \\
                            & POEM        &                      & \textbf{90.2} \\ \hline
\multirow{4}{*}{PointNet++} & Real-valued & 32/32                & 91.9          \\ \cline{2-4} 
                            & XNOR-Net     & \multirow{3}{*}{1/1} & 83.8          \\
                            & BiPointNet  &                      & 87.8          \\
                            & POEM        &                      & \textbf{91.2}          \\ \hline
\multirow{4}{*}{DGCNN}      & Real-valued & 32/32                & 89.2          \\ \cline{2-4} 
                            & XNOR-Net     & \multirow{3}{*}{1/1} & 81.5          \\
                            & BiPointNet  &                      & 83.4          \\
                            & POEM        &                      &        \textbf{91.1}       \\ \hline
\end{tabular}}
	\centering
	\vspace{3.3mm}
	\tabcaption{Our methods on mainstream networks on classification task with ModelNet40 dataset.}
	\label{modelnet}
	\end{minipage}
\end{figure}

We first introduce our baseline network by adding a single BN layer ahead of the 1-bit convolutions of XNOR-Net, which brings $\xu{1.2\%}$ improvement on OA. {As shown in Table~\ref{component}, the introduction of PReLU, EM, and learnable scale factor improves the accuracy by $1.9\%$, $3.1\%$ and $3.4\%$ respectively over the  baseline network, as shown in the second section of Table~\ref{component}.} By adding all the PReLU, EM, and the learnable scale factor, our POEM achieves $7.1\%$ higher accuracy than the baseline, even surpassing the corresponding real-valued network's accuracy. 

Compared to merely using the PReLU, our main contributions, EM and learnable scale factor, boost the accuracy by $5.2\%$, which is very significant on point cloud task. The 1-bit PointNet achieves the performance, which even surpasses the real-valued PointNet baseline with $1.0\%$ ($90.2\%$ vs. $89.2\%$). 

\noindent\textbf{Weight distribution:} The POEM-based model is based on an Expectation-Maximization process implemented on PyTorch \cite{paszke2019pytorch} platform. We analyze the weight distribution of training XNOR-Net and POEM for comparison to confirm our motivation. For a 1-bit PointNet model, we analyze the $6$-th 1-bit layer sized $(64, 64)$ and having $4096$ elements. We plot its weight distribution at the $\{0,40,60,120,160,200\}$-th epochs. As seen in Figure~\ref{distribution}, the initialization ($0$-th epoch) is the same for XNOR-Net and POEM. However, our POEM efficiently employs the Expectation-Maximization algorithm to supervise the back propagation process, leading to an effective and robust bi-modal distribution. This analysis also compiles with the performance comparison in Table~\ref{component}.

\subsection{Comparison with State-of-the-arts}

\noindent\textbf{Classification on ModelNet40:} 
Table~\ref{modelnet} shows that our POEM outperforms other binarization methods such as XNOR-Net \cite{rastegari2016xnor}, Bi-Real Net \cite{liu2018bi} and BiPointNet \cite{qin2020bipointnet} on classification task with ModelNet40 dataset. We implement comparative experiments on three mainstream backbones: PointNet \cite{qi2017pointnet}, PointNet++ \cite{qi2017pointnet++} and DGCNN \cite{wang2019dynamic}. XNOR-Net and Bi-Real Net have been proven effective in 2D vision, and we successfully transfer them to point clouds. 

Specifically, on PointNet, our POEM outperforms XNOR-Net, Bi-Real Net, and BiPointNet by $9.3\%$, $12.7\%$, and $3.8\%$ respectively. Moreover, 1-bit PointNet obtained by POEM even surpasses the real-valued PointNet by $1.0\%$ OA. On PointNet++, POEM stands out from all other 1-bit methods by a sizable performance advance. For example, POEM outperforms BiPointNet by $3.4\%$ OA improvement. Similar circumstances arise on the DGCNN backbone, and our POEM surpasses BiPointNet by $7.7\%$. All the results demonstrate that our POEM promotes the state-of-the-art in 1-bit point cloud classification.

\begin{figure}[t]
    \begin{minipage}[t]{0.48\textwidth}
    \vspace{-21.1mm}
    \footnotesize
    \centering
    \setlength{\tabcolsep}{1mm}{
\begin{tabular}{|c|c|c|c|}
\hline
Model               & Method      & W/A (bit)                & mIOU          \\ \hline
\multirow{5}{*}{PointNet}   & Real-valued & 32/32                & 83.7          \\ \cline{2-4} 
                            & XNOR-Net     & \multirow{4}{*}{1/1} & 75.3          \\
                            & Bi-Real Net &                      & 70.0          \\
                            & BiPointNet  &                      & 80.6          \\
                            & POEM        &                      & \textbf{81.1} \\ \hline
\multirow{3}{*}{PointNet++} & Real-valued & 32/32                & 85.1          \\ \cline{2-4} 
                            & XNOR-Net     & \multirow{2}{*}{1/1} & 77.7          \\
                            & POEM        &                      & \textbf{82.9} \\ \hline
\multirow{3}{*}{DGCNN}      & Real-valued & 32/32                & 85.2          \\ \cline{2-4} 
                            & XNOR-Net     & \multirow{2}{*}{1/1} & 77.4          \\
                            & POEM        &                      & \textbf{83.1}          \\ \hline
\end{tabular}}
    \vspace{1mm}
	\tabcaption{Our methods on mainstream networks on part segmentation task with ShapeNet Part dataset.}
	\label{part_seg}
	\end{minipage}
	\hspace{1mm}
	\begin{minipage}[t]{0.48\textwidth}
	\centering
	\footnotesize
\setlength{\tabcolsep}{0.1mm}{
\begin{tabular}{|c|c|c|c|c|}
\hline
Model                       & Method      & W/A (bit)                & mIOU          & OA (\%)            \\ \hline
\multirow{5}{*}{PointNet}   & Real-valued & 32/32                & 47.7          & 78.6          \\ \cline{2-5} 
                            & XNOR-Net     & \multirow{4}{*}{1/1} & 39.1          & 70.4          \\
                            & Bi-Real Net &                      & 35.5          & 65.0          \\
                            & BiPointNet  &                      & 44.3          & 76.7          \\
                            & POEM        &                      & \textbf{45.8} & \textbf{77.9} \\ \hline
\multirow{3}{*}{PointNet++} & Real-valued & 32/32                & 53.2          & 82.7          \\ \cline{2-5} 
                            & XNOR-Net     & \multirow{2}{*}{1/1} & 43.1          & 75.9          \\
                            & POEM        &                      & \textbf{49.8} & \textbf{80.4} \\ \hline
\multirow{3}{*}{DGCNN}      & Real-valued & 32/32                & 56.1          & 84.2          \\ \cline{2-5} 
                            & XNOR-Net     & \multirow{2}{*}{1/1} & 45.6          & 78.0          \\
                            & POEM        &                      & \textbf{50.1} & \textbf{81.3}          \\ \hline
\end{tabular}}
	\centering
	\vspace{1.7mm}
	\tabcaption{Our methods on mainstream networks on semantic segmentation task with S3DIS dataset.}
	\label{sem_seg}
	\end{minipage}
\end{figure}

\noindent\textbf{Part Segmentation on ShapeNet Parts:} We demonstrate the superiority of our POEM on part segmentation task in Table~\ref{part_seg}. We have two observations from the impressive results: 1). POEM can achieve the best performance (mIOU) compared with other 1-bit methods on all three backbones; 2). Compared with real-valued counterparts, acceptable performance drops are achieved ($2.6\%$, $2.2\%$ and $2.1\%$) with significant compression rates. 

\noindent\textbf{Semantic Segmentation on S3DIS:} As listed Table~\ref{sem_seg}, our POEM outperforms all other 1-bit methods on part segmentation tasks. First, POEM can achieve the best mIOU and OA compared with other 1-bit methods on all employed backbones. Second, POEM can inference the implemented backbones with efficient XNOR and Bit-count operations with acceptable mIOU drops ($1.9\%$, $3.4\%$ and $2.9\%$) achieved. Moreover, our POEM can achieve OA over $80\%$ with 1-bit weights and activations, which promotes the state-of-the-art. 

Except for these performance comparison, we also provide efficiency analysis and results visualization in the supplementary material to sufficiently evaluate our POEM.

\section{Conclusion}
We have developed a new deep learning model for point cloud processing, 1-bit point-wise operations based on Expectation-Maximization (POEM), which can significantly reduce the storage requirement for computationally limited devices. POEM is employed in point cloud networks to formulate 1-bit fully-connected layer (Bi-FC), which mainly works with binary weights, proposed scale factor, and Expectation-Maximization (EM) algorithm. In POEMs, we use the learnable scale factor to build an end-to-end framework and a new architecture to calculate the network model. To further enhance the robustness of unbinarized weights, we employ the EM algorithm to learning a bi-modal distribution of unbinarized weights. All the parameters of our POEM can be obtained in the same pipeline as in the back propagation algorithm. Extensive experiments demonstrate that our POEM surpasses existing binarization methods by significant margins. For more real-world applications, we will implement our POEM on ARM CPUs for future work.

\section{Acknowledgement}
Sheng Xu and Yanjing Li are the co-first authors. Baochang Zhang is the corresponding author. The work was supported by the 
National Natural Science Foundation of China (62076016, 61972016). This study was supported by Grant NO.2019JZZY011101 from the Key Research and Development Program of Shandong Province to Dianmin Sun.

\bibliographystyle{apalike}
\bibliography{bmvc_review}

\begin{thebibliography}{23}
\providecommand{\natexlab}[1]{#1}
\providecommand{\url}[1]{\texttt{#1}}
\expandafter\ifx\csname urlstyle\endcsname\relax
  \providecommand{\doi}[1]{doi: #1}\else
  \providecommand{\doi}{doi: \begingroup \urlstyle{rm}\Url}\fi

\bibitem[Armeni et~al.(2016)Armeni, Sener, Zamir, Jiang, Brilakis, Fischer, and
  Savarese]{armeni20163d}
Iro Armeni, Ozan Sener, Amir~R Zamir, Helen Jiang, Ioannis Brilakis, Martin
  Fischer, and Silvio Savarese.
\newblock 3d semantic parsing of large-scale indoor spaces.
\newblock In \emph{Proceedings of the IEEE Conference on Computer Vision and
  Pattern Recognition}, pages 1534--1543, 2016.

\bibitem[Chang et~al.(2015)Chang, Funkhouser, Guibas, Hanrahan, Huang, Li,
  Savarese, Savva, Song, Su, et~al.]{chang2015shapenet}
Angel~X Chang, Thomas Funkhouser, Leonidas Guibas, Pat Hanrahan, Qixing Huang,
  Zimo Li, Silvio Savarese, Manolis Savva, Shuran Song, Hao Su, et~al.
\newblock Shapenet: An information-rich 3d model repository.
\newblock \emph{arXiv preprint arXiv:1512.03012}, 2015.

\bibitem[Courbariaux et~al.(2015)Courbariaux, Bengio, and
  David]{courbariaux2015binaryconnect}
Matthieu Courbariaux, Yoshua Bengio, and Jean-Pierre David.
\newblock Binaryconnect: Training deep neural networks with binary weights
  during propagations.
\newblock In \emph{Advances in neural information processing systems}, pages
  3123--3131, 2015.

\bibitem[Gu et~al.({\natexlab{a}})Gu, Li, Zhang, Han, Cao, Liu, and
  Doermann]{gu2019projection}
Jiaxin Gu, Ce~Li, Baochang Zhang, Jungong Han, Xianbin Cao, Jianzhuang Liu, and
  David Doermann.
\newblock Projection convolutional neural networks for 1-bit cnns via discrete
  back propagation.
\newblock In \emph{Proceedings of AAAI Conference on Artificial Intelligence},
  pages 8344--8351, {\natexlab{a}}.

\bibitem[Gu et~al.({\natexlab{b}})Gu, Zhao, Jiang, Zhang, Liu, Guo, and
  Ji]{gu2019bayesian}
Jiaxin Gu, Junhe Zhao, Xiaolong Jiang, Baochang Zhang, Jianzhuang Liu, Guodong
  Guo, and Rongrong Ji.
\newblock Bayesian optimized 1-bit cnns.
\newblock In \emph{Proceedings of IEEE International Conference on Computer
  Vision}, pages 4909--4917, {\natexlab{b}}.

\bibitem[He et~al.(2015)He, Zhang, Ren, and Sun]{he2015delving}
Kaiming He, Xiangyu Zhang, Shaoqing Ren, and Jian Sun.
\newblock Delving deep into rectifiers: Surpassing human-level performance on
  imagenet classification.
\newblock In \emph{Proceedings of the IEEE International Conference on Computer
  Vision}, pages 1026--1034, 2015.

\bibitem[Hermosilla et~al.(2019)Hermosilla, Ritschel, and
  Ropinski]{hermosilla2019total}
Pedro Hermosilla, Tobias Ritschel, and Timo Ropinski.
\newblock Total denoising: Unsupervised learning of 3d point cloud cleaning.
\newblock In \emph{Proceedings of the IEEE International Conference on Computer
  Vision}, pages 52--60, 2019.

\bibitem[Hu et~al.(2020)Hu, Yang, Xie, Rosa, Guo, Wang, Trigoni, and
  Markham]{hu2020randla}
Qingyong Hu, Bo~Yang, Linhai Xie, Stefano Rosa, Yulan Guo, Zhihua Wang, Niki
  Trigoni, and Andrew Markham.
\newblock Randla-net: Efficient semantic segmentation of large-scale point
  clouds.
\newblock In \emph{Proceedings of the IEEE Conference on Computer Vision and
  Pattern Recognition}, pages 11108--11117, 2020.

\bibitem[Liu et~al.(2018)Liu, Wu, Luo, Yang, Liu, and Cheng]{liu2018bi}
Zechun Liu, Baoyuan Wu, Wenhan Luo, Xin Yang, Wei Liu, and Kwang-Ting Cheng.
\newblock Bi-real net: Enhancing the performance of 1-bit cnns with improved
  representational capability and advanced training algorithm.
\newblock In \emph{Proceedings of European Conference on Computer Vision},
  pages 722--737, 2018.

\bibitem[Liu et~al.(2020)Liu, Shen, Savvides, and Cheng]{liu2020reactnet}
Zechun Liu, Zhiqiang Shen, Marios Savvides, and Kwang-Ting Cheng.
\newblock Reactnet: Towards precise binary neural network with generalized
  activation functions.
\newblock In \emph{Proceedings of European Conference on Computer Vision},
  pages 143--159, 2020.

\bibitem[Liu et~al.(2019)Liu, Tang, Lin, and Han]{liu2019point}
Zhijian Liu, Haotian Tang, Yujun Lin, and Song Han.
\newblock Point-voxel cnn for efficient 3d deep learning.
\newblock In \emph{Proceedings of Advances in Neural Information Processing
  Systems}, pages 965--975, 2019.

\bibitem[Moon(1996)]{moon1996expectation}
Todd~K Moon.
\newblock The expectation-maximization algorithm.
\newblock \emph{IEEE Signal processing magazine}, 13\penalty0 (6):\penalty0
  47--60, 1996.

\bibitem[Paszke et~al.(2019)Paszke, Gross, Massa, Lerer, Bradbury, Chanan,
  Killeen, Lin, Gimelshein, Antiga, et~al.]{paszke2019pytorch}
Adam Paszke, Sam Gross, Francisco Massa, Adam Lerer, James Bradbury, Gregory
  Chanan, Trevor Killeen, Zeming Lin, Natalia Gimelshein, Luca Antiga, et~al.
\newblock Pytorch: An imperative style, high-performance deep learning library.
\newblock In \emph{Advances in Neural Information Processing Systems}, pages
  8026--8037, 2019.

\bibitem[Qi et~al.(2017{\natexlab{a}})Qi, Su, Mo, and Guibas]{qi2017pointnet}
Charles~R Qi, Hao Su, Kaichun Mo, and Leonidas~J Guibas.
\newblock Pointnet: Deep learning on point sets for 3d classification and
  segmentation.
\newblock In \emph{Proceedings of the IEEE Conference on Computer Vision and
  Pattern Recognition}, pages 652--660, 2017{\natexlab{a}}.

\bibitem[Qi et~al.(2017{\natexlab{b}})Qi, Yi, Su, and Guibas]{qi2017pointnet++}
Charles~Ruizhongtai Qi, Li~Yi, Hao Su, and Leonidas~J Guibas.
\newblock Pointnet++: Deep hierarchical feature learning on point sets in a
  metric space.
\newblock In \emph{Proceedings of Advances in Neural Information Processing
  Systems}, pages 5099--5108, 2017{\natexlab{b}}.

\bibitem[Qin et~al.(2020)Qin, Cai, Zhang, Ding, Zhao, Yi, Liu, and
  Su]{qin2020bipointnet}
Haotong Qin, Zhongang Cai, Mingyuan Zhang, Yifu Ding, Haiyu Zhao, Shuai Yi,
  Xianglong Liu, and Hao Su.
\newblock Bipointnet: Binary neural network for point clouds.
\newblock In \emph{Proceedings of International Conference on Learning
  Representations}, pages 1--24, 2020.

\bibitem[Rastegari et~al.(2016)Rastegari, Ordonez, Redmon, and
  Farhadi]{rastegari2016xnor}
Mohammad Rastegari, Vicente Ordonez, Joseph Redmon, and Ali Farhadi.
\newblock Xnor-net: Imagenet classification using binary convolutional neural
  networks.
\newblock In \emph{Proceedings of European Conference on Computer Vision},
  pages 525--542, 2016.

\bibitem[Wang et~al.(2019)Wang, Sun, Liu, Sarma, Bronstein, and
  Solomon]{wang2019dynamic}
Yue Wang, Yongbin Sun, Ziwei Liu, Sanjay~E Sarma, Michael~M Bronstein, and
  Justin~M Solomon.
\newblock Dynamic graph cnn for learning on point clouds.
\newblock \emph{Acm Transactions On Graphics}, 38\penalty0 (5):\penalty0 1--12,
  2019.

\bibitem[Wang et~al.(2020)Wang, Wu, Lu, and Zhou]{wang2020bidet}
Ziwei Wang, Ziyi Wu, Jiwen Lu, and Jie Zhou.
\newblock Bidet: An efficient binarized object detector.
\newblock In \emph{Proceedings of IEEE Conference on Computer Vision and
  Pattern Recognition}, pages 2049--2058, 2020.

\bibitem[Wu et~al.(2015)Wu, Song, Khosla, Yu, Zhang, Tang, and Xiao]{wu20153d}
Zhirong Wu, Shuran Song, Aditya Khosla, Fisher Yu, Linguang Zhang, Xiaoou Tang,
  and Jianxiong Xiao.
\newblock 3d shapenets: A deep representation for volumetric shapes.
\newblock In \emph{Proceedings of the IEEE Conference on Computer Vision and
  Pattern Recognition}, pages 1912--1920, 2015.

\bibitem[Xu et~al.(2020{\natexlab{a}})Xu, Sun, Wu, Wang, and
  Neumann]{xu2020grid}
Qiangeng Xu, Xudong Sun, Cho-Ying Wu, Panqu Wang, and Ulrich Neumann.
\newblock Grid-gcn for fast and scalable point cloud learning.
\newblock In \emph{Proceedings of the IEEE Conference on Computer Vision and
  Pattern Recognition}, pages 5661--5670, 2020{\natexlab{a}}.

\bibitem[Xu et~al.(2020{\natexlab{b}})Xu, Liu, Gong, Liu, Mao, and
  Zhang]{xu2020amplitude}
Sheng Xu, Zhendong Liu, Xuan Gong, Chunlei Liu, Mingyuan Mao, and Baochang
  Zhang.
\newblock Amplitude suppression and direction activation in networks for 1-bit
  faster r-cnn.
\newblock In \emph{Proceedings of the International Workshop on Embedded and
  Mobile Deep Learning}, pages 19--24, 2020{\natexlab{b}}.

\bibitem[Xu et~al.(2021)Xu, Zhao, Lu, Zhang, Han, and Doermann]{xu2020layer}
Sheng Xu, Junhe Zhao, Jinhu Lu, Baochang Zhang, Shumin Han, and David Doermann.
\newblock Layer-wise searching for 1-bit detectors.
\newblock In \emph{Proceedings of the IEEE Conference on Computer Vision and
  Pattern Recognition}, pages 5682--5691, 2021.

\end{thebibliography}

\end{document}